\documentclass{article}

\PassOptionsToPackage{numbers, sort&compress}{natbib}
\usepackage[final, dblblindworkshop]{neurips_2026}
\workshoptitle{Physical World AI: Geometry, Characteristics, and Multimodal Sensing}

\usepackage[utf8]{inputenc} % allow utf-8 input
\usepackage[T1]{fontenc}    % use 8-bit T1 fonts
\usepackage{hyperref}       % hyperlinks
\hypersetup{hidelinks}      % links stay clickable, no coloured boxes
\usepackage{url}            % simple URL typesetting
\usepackage{booktabs}       % professional-quality tables
\usepackage{amsfonts}       % blackboard math symbols
\usepackage{nicefrac}       % compact symbols for 1/2, etc.
\usepackage{microtype}      % microtypography
\usepackage{xcolor}         % colors
\usepackage{multirow}
\usepackage{amsmath}
\usepackage{graphicx}

\title{
Do 3D Point Representations Encode Architectural Functions?\\
Probing Supervised and Pretrained Models on Urban Facades
}

\author{%
  Abdiaziz Muse \quad Jiju Poovvancheri \\
  Graphics and Spatial Computing Lab, Saint Mary's University \\
  Halifax, Canada \\
  \texttt{\{abdiaziz.muse, jiju.poovvancheri\}@smu.ca}
}

\begin{document}

\maketitle

\begin{abstract}
Semantic labels say what a building element \emph{is}, not what it
\emph{does}: two identical-looking windows on one facade may be operable or
fixed, a column may carry load or only decorate. We ask whether 3D point
representations encode such architectural function. On 22 mobile-LiDAR
buildings with 15 functional attributes per element, we hold out entire
buildings and compare hand-crafted descriptors, networks trained from
scratch and frozen self-supervised point transformers. Descriptors and
frozen features are read by the same linear layer. No
network trained from scratch, from PointNet to a 108M-parameter PTv3,
clearly exceeds the descriptors (0.610 mAP on functionally ambiguous
elements), whereas frozen Utonia reaches 0.732. Random-weight controls
of the same architectures (0.649 and 0.652) attribute the gain to
pretraining (Utonia: +0.080, 95\% building interval $[+0.035, +0.124]$). Pretrained features improve at every tested number
of labelled buildings while descriptors and random features plateau, and, in the configurations we tested, fine-tuning erodes the benefit: at today's data scale, function is better read from pretrained representations than learned from scratch.
\end{abstract}

% ============================================================
\section{Introduction}
% ============================================================

\begin{figure}[t]
    \centering
    \includegraphics[width=\linewidth]{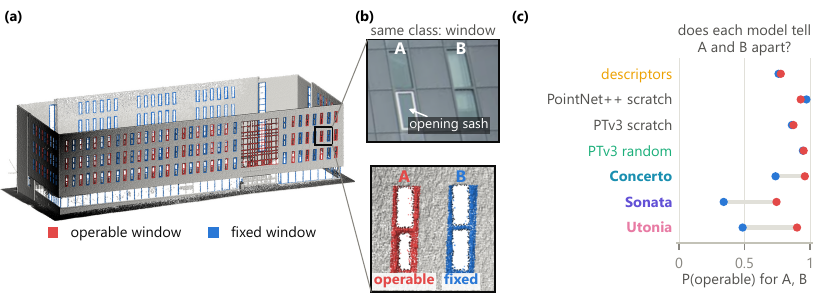}
    \caption{
    \textbf{Same class, different function.}
    (a)~Held-out building; windows coloured by ground-truth operability.
    (b)~Windows A and B. Top: Google Street View (\copyright{}\,Google), used to verify
    the labels; A has an opening sash. Bottom: the same windows in the LiDAR scan.
    (c)~$P(\text{operable})$: the predicted probability that A (red) and B (blue)
    are operable, from a linear layer on each feature source.
    }
    \label{fig:teaser}
    \vspace{-15pt}
\end{figure}

3D scene understanding has made steady progress on \emph{what} is present
and \emph{where}~\cite{qi2017pointnet,wu2024ptv3,hu2021sensaturban};
physical-world reasoning also needs \emph{what an object
does}~\cite{deng2021affordancenet,delitzas2024scenefun3d}. On a facade, function governs operation, safety and energy use:
robots, inspection and retrofit tools need to know whether a window opens,
a column carries load or a stair edge protects against falls. These
properties are not fixed by the semantic class~\cite{hfx3d}: two windows side by side
may be operable and fixed (Fig.~\ref{fig:teaser}a,b) yet look nearly
identical in a LiDAR scan, since glass returns few points and an opening
sash is a thin frame.

Learning function directly is hard because labels are scarce: each building
contributes one style and largely one set of functional decisions. Moreover, to our knowledge, no existing network is designed specifically for instance-level functional attribute prediction from 3D facade point clouds. With
$\sim$18 training buildings, we find that point networks trained from
scratch, including a purpose-built model fusing shape, semantics and
relational context, do not clearly exceed 21 hand-crafted descriptors.
Self-supervised point backbones pretrained on tens to hundreds of thousands
of unlabelled point clouds~\cite{wu2025sonata,zhang2025concerto,zhang2026utonia}, however, are
known to encode semantics and geometry. We therefore ask:

\begin{quote}
\emph{Do frozen pretrained 3D representations encode architectural function, and is this functional signal acquired through pretraining rather than provided by the network architecture alone?}
\end{quote}

We study this question through a controlled probing framework~\cite{alain2017probes}: each method receives the same facade element points and semantic class, and function is predicted using the same linear probe; only the underlying feature representation changes. Our contributions are threefold.
(i)~a probing protocol for architectural function on real outdoor LiDAR,
with whole buildings held out and a random-weight control that separates
pretraining from architecture;
(ii)~evidence that frozen pretrained features encode facade function that
our networks did not learn from $\sim$18 labelled buildings, with the gain attributable to
pretraining, growing with more labels, and eroded by fine-tuning;
(iii)~an analysis of which functions benefit, with the largest gains observed for window operability
and ventilation.

% ============================================================
\section{Related Work}
% ============================================================

\paragraph{Function in 3D.}
Functionality understanding has moved from object-level affordance
labels~\cite{deng2021affordancenet,li2024laso} to scene-level benchmarks of
interactive elements~\cite{delitzas2024scenefun3d}, addressed with
foundation-model pipelines~\cite{corsetti2025fun3du} and functional scene
graphs~\cite{zhang2025openfungraph,fu2026funfact}, all indoors or on
objects. Urban, heritage and facade
point-cloud benchmarks annotate semantic
classes~\cite{hu2021sensaturban,koelle2021h3d,tan2020toronto3d,matrone2020arch,wysocki2022tumfacade,wysocki2025zaha},
mainly for segmentation and LoD3 reconstruction~\cite{wysocki2023scan2lod3},
and image-based work adds building
attributes~\cite{liang2025openfacades}. To our knowledge, whether 3D point
representations encode the \emph{function} of building elements is untested.

\paragraph{Point pretraining and probing.}
Self-supervised point pretraining has progressed from object-level masked
modelling~\cite{yu2022pointbert,pang2022pointmae} and scene
contrast~\cite{xie2020pointcontrast,wu2023msc,hou2021csc} to large
multi-dataset PTv3 encoders~\cite{wu2024ptv3,wu2024ppt} whose frozen features
support linear read-out~\cite{wu2025sonata,zhang2025concerto,zhang2026utonia}.
We follow the linear-probing tradition~\cite{alain2017probes}; frozen-feature
probes have measured the 3D awareness of foundation models~\cite{elbanani2024probe3d,man2024lexicon3d}.
Concurrent work pretrains a city-scale encoder and compares it with these
checkpoints under frozen probes for urban and facade
\emph{segmentation}~\cite{rusnak2026polis}; we probe \emph{function},
holding out whole buildings.

% ============================================================
\section{Task and Evaluation}
\label{sec:task}
% ============================================================

\paragraph{Data and task.}
We use the HFX3D benchmark~\cite{hfx3d}, which contains mobile-LiDAR scans of real
buildings in which every facade element (e.g.\ window, wall, column, arch, roof,
stairs) carries a label vector $\mathbf{y}_i\in\{0,1\}^{15}$ over
\emph{access, aesthetic, circulation, drainage, fall protection,
illumination, load bearing, natural lighting, operability, privacy
screening, solar shading, surveillance, thermal envelope, vegetation
support} and \emph{ventilation}, validated by annotators against
street-level imagery. We use its 22 development buildings (5,172 elements).

\paragraph{Functionally ambiguous pairs.}
Many attributes are fixed by the class (every window admits natural light).
Following~\cite{hfx3d}, a class--attribute pair $(c,a)$ is \emph{ambiguous}
when its training frequency
$r(c,a)=|\{i:c_i=c,\,y_{i,a}=1\}|/|\{i:c_i=c\}|$ satisfies
$0.05\le r(c,a)\le 0.95$. This yields 16 pairs over seven classes, e.g.\
window operability, column load bearing, roof drainage and stair fall
protection. All results are on the 2,472 elements of these seven classes.

\paragraph{Protocol.}
Functional labels, geometry, and style correlate within buildings, making random element-level splits susceptible to building-specific leakage. We use five building-disjoint folds over 22 buildings, holding out 3--5 per fold and training on the remaining $\sim$18. Every configuration is run
with 10 seeds and a fixed schedule, without checkpoint selection. We report mAP over 12 attributes with both positive and negative test examples; AP pools eligible instances from seven semantic classes, and mAP averages attributes equally. mAP is threshold-free and measures whether elements with a
function rank above those without. A random ranking scores, for each
attribute, the share of elements that have the function; averaged over the
12 attributes, this chance level is 0.569. A useful model must score above it. We compute 95\% confidence intervals from 2,000 building-level bootstrap resamples paired across models; $\Delta$ is the mean per-seed difference.

% ============================================================
\section{Probing Setup}
\label{sec:setup}
% ============================================================

\textbf{Hand-crafted features.} We compute 21 metric descriptors per
element, such as bounding-box size, height above ground, planarity and
verticality.
\textbf{Networks trained from scratch.} We train PointNet~\cite{qi2017pointnet},
PointNet++~\cite{qi2017pointnet2} and DGCNN~\cite{wang2019dgcnn}, each with
about 1M parameters and 2,048 input points per element, and PTv3~\cite{wu2024ptv3}
in Sonata's 108M configuration~\cite{wu2025sonata} end to end on the benchmark labels.
The point networks train for 500 epochs. PTv3 trains for 3,000 steps, each
on a random 9\,m window: a 6\,m core with 1.5\,m of context, the same
tiling as for the frozen features.
\textbf{Random-weight controls.} We freeze each pretrained architecture at
its random initialisation. Untrained networks, both
random-weight~\cite{saxe2011random} and non-parametric~\cite{zhang2023pointnn},
can already turn local structure into informative features. These controls
therefore measure what the architecture and spatial context contribute
without pretraining.
\textbf{Frozen pretrained models.} We use the public self-supervised
checkpoints of Sonata~\cite{wu2025sonata}, Concerto~\cite{zhang2025concerto}
and Utonia~\cite{zhang2026utonia}. None of them was trained on HFX3D.

For the frozen backbones, each building is tiled into 6\,m cubes with a
1.5\,m margin of context, voxelised at 2\,cm with LiDAR intensity and
estimated normals, and encoded once; point features are mean- and
max-pooled over each element, giving $\mathbf{f}_i\in\mathbb{R}^{2D}$.
The frozen backbones thus see each element with its surroundings, while
the from-scratch networks see its points alone; the random-weight control
receives exactly the frozen backbones' input. The probe is
$\hat{\mathbf{y}}_i=\sigma\big(W[\mathbf{f}_i;\mathrm{onehot}(c_i)]+\mathbf{b}\big)$,
trained with binary cross-entropy on applicable attributes for 400 epochs. The from-scratch encoders end in the same linear layer
over their pooled embedding and a class embedding; PTv3 trained from
scratch predicts per point and averages over the element. We also report a
three-branch model we designed for the task, combining shape, descriptors,
class and a neighbour graph.

% ============================================================
\section{Results}
\label{sec:results}
% ============================================================

\begin{table}[t]
\centering
\caption{
\textbf{Function probing on held-out buildings.} mAP (mean $\pm$ sd over 10
seeds; 5 for DGCNN); $\Delta$: vs.\ random-weight PTv3 (Utonia: vs.\ random Utonia), with 95\% building interval. Chance: 0.569.
}
\label{tab:main}
\footnotesize
\setlength{\tabcolsep}{3pt}
\begin{tabular}{lllllcc}
\toprule
Features & Network & Venue & Pretraining data & Trained & mAP & $\Delta$ vs.\ random \\
\midrule
Class only & one-hot & -- & -- & linear & 0.556\,$\pm$\,.011 & $-0.093$ \scriptsize{[$-.144$, $-.022$]} \\
Hand-crafted & 21 descriptors & -- & -- & linear & 0.610\,$\pm$\,.004 & $-0.040$ \scriptsize{[$-.079$, $+.017$]} \\
\midrule
\multirow{4}{*}{From scratch}
 & PointNet~\cite{qi2017pointnet} & CVPR'17 & -- & all ($\sim$1M) & 0.612\,$\pm$\,.018 & $-0.037$ \scriptsize{[$-.073$, $+.017$]} \\
 & PointNet++~\cite{qi2017pointnet2} & NeurIPS'17 & -- & all ($\sim$1M) & 0.622\,$\pm$\,.014 & $-0.027$ \scriptsize{[$-.086$, $+.035$]} \\
 & DGCNN~\cite{wang2019dgcnn} & TOG'19 & -- & all ($\sim$1M) & 0.613\,$\pm$\,.014 & $-0.038$ \scriptsize{[$-.083$, $+.017$]} \\
 & PTv3~\cite{wu2024ptv3} & CVPR'24 & -- & all (108M) & 0.548\,$\pm$\,.012 & $-0.101$ \scriptsize{[$-.158$, $-.035$]} \\
\midrule
\multirow{2}{*}{Random, frozen}
 & PTv3~\cite{wu2024ptv3} & CVPR'24 & none & linear & 0.649\,$\pm$\,.002 & reference \\
 & Utonia~\cite{zhang2026utonia} & ICML'26 & none & linear & 0.652\,$\pm$\,.002 & reference \\
\midrule
\multirow{3}{*}{\shortstack[l]{Pretrained,\\frozen}}
 & Concerto~\cite{zhang2025concerto} & NeurIPS'25 & indoor + images & linear & 0.694\,$\pm$\,.002 & $+0.044$ \scriptsize{[$-.002$, $+.093$]} \\
 & Sonata~\cite{wu2025sonata} & CVPR'25 & indoor        & linear & 0.716\,$\pm$\,.004 & $+0.067$ \scriptsize{[$-.008$, $+.112$]} \\
 & Utonia~\cite{zhang2026utonia} & ICML'26 & multi-domain & linear & \textbf{0.732}\,$\pm$\,.005 & $\mathbf{+0.080}$ \scriptsize{[$+.035$, $+.124$]} \\
\bottomrule
\end{tabular}
\vspace{-18pt}
\end{table}

\paragraph{Training from scratch does not beat descriptors.}
We train PointNet~\cite{qi2017pointnet}, PointNet++~\cite{qi2017pointnet2},
DGCNN~\cite{wang2019dgcnn} and PTv3~\cite{wu2024ptv3} from scratch with a
standard recipe and no per-model tuning. From $\sim$1M-parameter point networks to the
108M-parameter PTv3, all land at 0.548--0.628 mAP; none, including the
three-branch model with descriptors and relational context (0.628), is reliably
better than the hand-crafted descriptors (Table~\ref{tab:main}). A linear layer
seeing only each element's class scores 0.556, below chance: it gives every
element of a class the same score and cannot rank one window above another. PTv3
trained from scratch even falls below its random-weight version ($-0.101$)
and below chance, as it overfits the $\sim$18 training buildings. With a
positive-weighted loss, the probes reach 0.662 on descriptors and 0.656 on
random weights, while Concerto (0.708), Sonata (0.724) and Utonia (0.733)
stay ahead.

\begin{figure}[t]
\begin{minipage}[c]{0.49\linewidth}
    \centering
    \includegraphics[width=\linewidth]{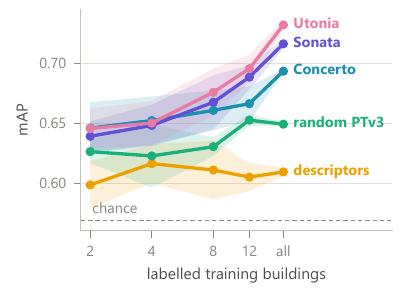}
\end{minipage}\hfill
\begin{minipage}[c]{0.49\linewidth}
    \centering
    \footnotesize
    \setlength{\tabcolsep}{3pt}
    \begin{tabular}{lc}
    \toprule
    \emph{PTv3, per-point head, no class} & mAP \\
    \midrule
    random init., trained & 0.548 \\
    Sonata init., fine-tuned & 0.559 \\
    Sonata init., fine-tuned, lr\,/\,50 & 0.658 \\
    \textbf{Sonata init., frozen} & \textbf{0.678} \\
    \midrule
    \emph{Linear probe (pooled + class)} & \\
    Sonata / +\,21 descriptors & 0.716 / 0.716 \\
    Utonia / +\,21 descriptors & 0.732 / 0.735 \\
    Sonata context: none / 1.5\,m / 3\,m & 0.685 / \textbf{0.716} / 0.714 \\
    \bottomrule
    \end{tabular}
\end{minipage}
\caption{
\textbf{Left:} mAP vs.\ number of labelled training buildings (5 draws;
all: $\sim$18; band: $\pm$ sd). \textbf{Right:} fine-tuning, added descriptors and context
(mAP).
}
\label{fig:curve}
\vspace{-15pt}
\end{figure}

\paragraph{Pretraining is what helps.}
In Table~\ref{tab:main}, we compare each pretrained model with its own
architecture at random weights. Pretraining adds up to $+0.080$ (Utonia),
the only $\Delta$ whose interval excludes zero; Utonia's margin over Sonata
($+0.016$) is within uncertainty. With the same per-point head (Fig.~\ref{fig:curve}, right),
frozen Sonata exceeds PTv3 trained from scratch by $+0.130$ $[+0.089, +0.188]$.

\paragraph{Pretrained features keep learning from labels.}
As the probe sees more labelled buildings (Fig.~\ref{fig:curve}, left),
pretrained features improve at every step, while descriptors and random
weights saturate early. With two buildings, Utonia (0.646) already exceeds
descriptors trained on all of them (0.610).

\paragraph{Freezing beats our fine-tuning; descriptors are redundant.}
Fine-tuning Sonata end to end destroys the benefit (0.559), and a
50$\times$ lower backbone learning rate (0.658) stays below the frozen
backbone (0.678), though within uncertainty and specific to the fine-tuning
schedules we tried (Fig.~\ref{fig:curve}, right).
The 21 descriptors add no measurable information to frozen features
($+0.003$ at most). Context helps up to 1.5\,m and no further; even without
context, Sonata (0.685) exceeds every network trained from scratch.

\paragraph{Which functions benefit.}
We score each ambiguous pair by its AP lift over chance,
$(\mathrm{AP}-p)/(1-p)$, where $p$ is the pair's positive rate.
Relative to its random-weight control, Utonia raises this lift most for window operability ($+0.58$)
and ventilation ($+0.49$), with further gains for column aesthetic ($+0.45$)
and wall vegetation support ($+0.34$); no representation ranks solar shading.
The gain is not only building recognition. Within held-out buildings,
Utonia ranks windows by operability with AUC 0.78, against 0.48 for random
weights; over all 14 pairs that vary inside a building, its mean AUC is
0.81, against 0.51--0.56. Fig.~\ref{fig:teaser}c shows one pair,
chosen before inspecting predictions.

% ============================================================
\section{Limitations and Conclusion}
% ============================================================

Semantic labels describe what architectural elements are; their function is
already partly encoded in pretrained point representations.
At today's scale of annotated building data, reading function from a frozen backbone outperforms training from scratch. This gain is primarily attributable to pretraining rather than architecture, particularly for Utonia, whose improvement over its random-weight control is the only one with an interval that excludes zero. Whether outdoor pretraining data helps further
needs more buildings to tell. Our study is limited to 22 buildings from
one city; Sonata's and Concerto's gains over the random control are not
significant at the building level, and a linear probe gives a lower bound on
what the representations encode; a stronger read-out may extract more.
Comparisons also differ in spatial context and training setup: the frozen
backbones encode each element with its surroundings, whereas the networks
trained from scratch see its points alone.

{
\small
\bibliographystyle{plainnat}
\bibliography{refs}
}

%%%%%%%%%%%%%%%%%%%%%%%%%%%%%%%%%%%%%%%%%%%%%%%%%%%%%%%%%%%%

\end{document}